\documentclass[10pt]{article}
\usepackage{2022abstract}
\usepackage{booktabs}
\usepackage{xcolor}
\usepackage{subcaption}
\usepackage{tikz}
\usepackage[percent]{overpic}
\usepackage{arydshln} 
\usepackage{graphicx}
\usepackage{amsmath}
\usepackage{adjustbox}
\usepackage{helvet} 

\title{Large Language Models Show Human-like Social Desirability Biases in Survey Responses} 

\author[1]{Aadesh Salecha}
\author[2]{Molly E. Ireland}
\author[1]{Shashanka Subrahmanya}
\author[3]{João Sedoc}
\author[4]{Lyle H. Ungar}
\author[1]{Johannes C. Eichstaedt}
\affil[1]{Stanford University}
\affil[2]{Receptiviti}
\affil[3]{New York University}
\affil[4]{University of Pennsylvania}

\begin{document} 
\maketitle
\begin{multicols}{2}

\section{Abstract}
Large Language Models (LLMs) are becoming more widely used to simulate human participants and so 
understanding their biases is important. We developed an experimental framework using Big Five personality surveys and uncovered a previously undetected social desirability bias in a wide range of LLMs. 
By systematically varying the number of questions LLMs were exposed to, we demonstrate their ability to infer when they are being evaluated. 
When personality evaluation is inferred, LLMs skew
their scores towards the desirable ends of trait dimensions (i.e., increased extraversion, decreased neuroticism, etc). This bias exists in all tested models, including GPT-4/3.5, Claude 3, Llama 3, and PaLM-2. 
Bias levels 
appear to increase in more recent models, 
with GPT-4's survey responses changing by 1.20 (human) standard deviations and Llama 3's by 0.98 standard deviations, which are very large effects.
This bias remains after question order randomization and paraphrasing. Reverse-coding the questions decreases bias levels but does not eliminate them, suggesting that this effect cannot be attributed to acquiescence bias. 
Our findings reveal an emergent social desirability bias and suggest constraints 
on profiling LLMs with psychometric tests and  
on this use of LLMs as proxies for human participants.

\vspace{2mm}
\noindent\Keywords{large language models}{cognitive bias}{AI}{psychometrics}{emergent behavior}\\

{\Large{L}}arge Language Models have demonstrated remarkable proficiency in a wide array of tasks, ranging from language translation and creative writing to code generation problem-solving. 
As these models are trained on vast amounts of human-generated text data, they can emulate human textual behavior and exhibit emergent capabilities that were not anticipated during their development~\cite{wei2022emergent}. Researchers are using LLMs' emulation capabilities 
to simulate data from human participants across diverse psychological and behavioral experiments~\cite{aher2023using,argyle2023out,dillion2023can}. 

Personality traits like the Big Five are among the most studied individual differences among human subjects.
The Big Five aim to provide a concise summary of between-person trait differences, spanning Extraversion, Openness to Experience, Conscientiousness, Agreeableness, and Neuroticism. Though originally intended to be value-neutral, most Big Five assessments are evaluative~\cite{wood2022less}, with people preferring lower Neuroticism (i.e., negativity and vulnerability to stress) and higher scores on the remaining four traits. Big Five traits are also robust predictors of various human behavioral tendencies and life outcomes\cite{roberts2007power}. 
Researchers have assessed LLMs with Big Five surveys and compared trait distribution to human norms~\cite{mei2024turing}, in addition to correlating traits with downstream task performance~\cite{safdari2023personality}. 
%
However,But the assessment of these personality traits via self-report questionnaires is vulnerable to response biases~\cite{schwarz1999self} such as acquiescence (tending to agree with questions regardless of their content) and social desirability biases (skewing responses toward perceived societal ideals). Prior research has used psychology experiments with LLMs~\cite{binz2023using} to document acquiescence and other biases~\cite{acquiescne}. Social desirability biases are some of the most persistent sources of error variance on surveys relevant to social norms, such as most personality traits~\cite{holtgraves2004social}. 


To evaluate response biases in LLMs, we conducted a series of experiments using a standardized 100-item Big Five personality questionnaire. We administered the questionnaire in batches, systematically varying the number of questions per batch (denoted as $Q_n$). To ensure the LLM had no access to previous items, we started a new context window (session) for each batch. We provided standardized instructions asking the LLMs to respond on a 5-point Likert scale. We analyzed models from OpenAI, Anthropic, Google, and Meta to ensure broad generalizability.  By systematically varying the number of questions, we uncovered a persistent social desirability bias across LLMs. See Supplementary Information (SI) for full methods.

\section*{Results}

\subsection{Manifestation of Social Desirability Bias}
\begin{figure}[H] 
\centering
\raisebox{-.5\height}{
  \begin{overpic}[height=115px]{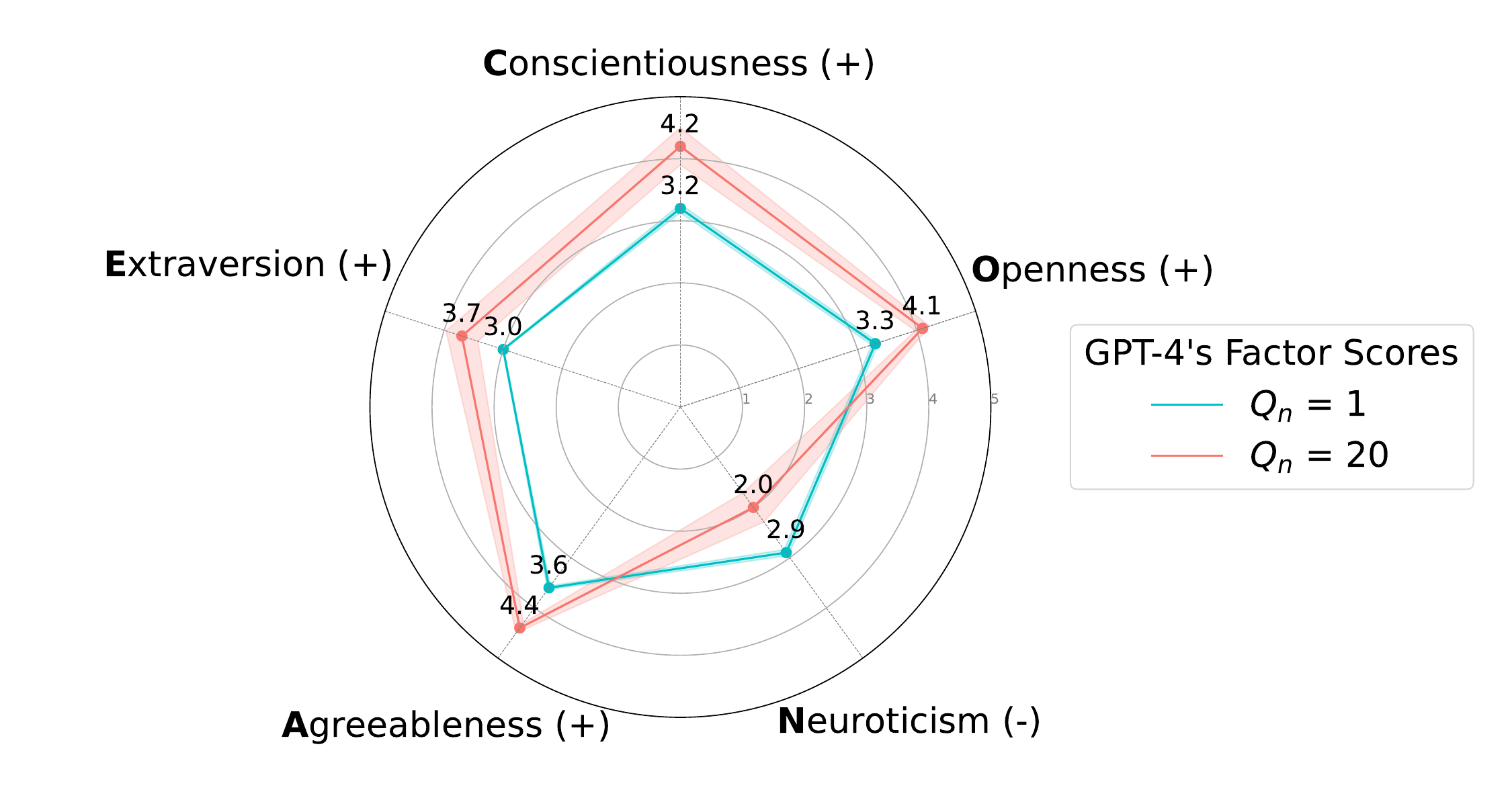}
    \put(-6,115px){\color{black}\textsf{\textbf{A}}} 
  \end{overpic}
}\\
\vspace{0.5mm}
{\raggedright \textsf{\textbf{B}} \par} 
\vspace{1mm}
{
\footnotesize
\begin{tabular}{p{70px}p{15px}p{15px}p{30px}p{50px}}
\toprule
\textbf{Trait} & \pmb{$Q_{n}$} \textbf{= 1} & \pmb{$Q_{n}$} \textbf{= 20} & \textbf{GPT-4 Diff.} & \textbf{Diff. in units of human SD~\cite{hughes2021big}} \\
\midrule
Openness (+) & 3.31 & 4.09 & +0.77 & \textbf{1.40} \\
Conscientiousness (+) & 3.25 & 4.20 & +0.96 & \textbf{1.42} \\
Extroversion (+) & 3.00 & 3.74 & +0.73 & \textbf{0.89} \\
Agreeableness (+) & 3.64 & 4.41 & +0.77 & \textbf{1.20} \\
Neuroticism (+) & 2.87 & 2.02 & -0.85 & \textbf{1.11} \\
\midrule
\textbf{Average} &  &  & \-\ 0.82 & \textbf{1.20} \\
\bottomrule
\vspace{15mm}
\end{tabular}
}
\raisebox{-.5\height}{\begin{overpic}[width=\linewidth]{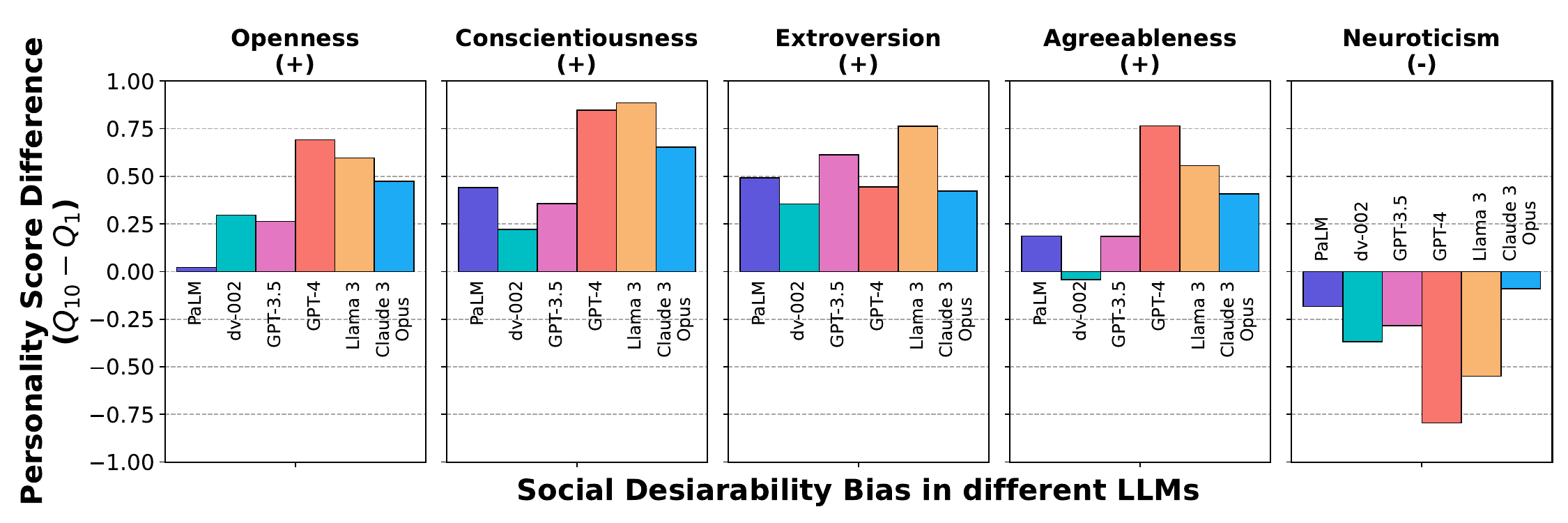}
  \put(0,82px){\color{black}\textsf{\textbf{C}}}
\end{overpic}}
\raisebox{-.5\height}{\begin{overpic}[width=\linewidth]{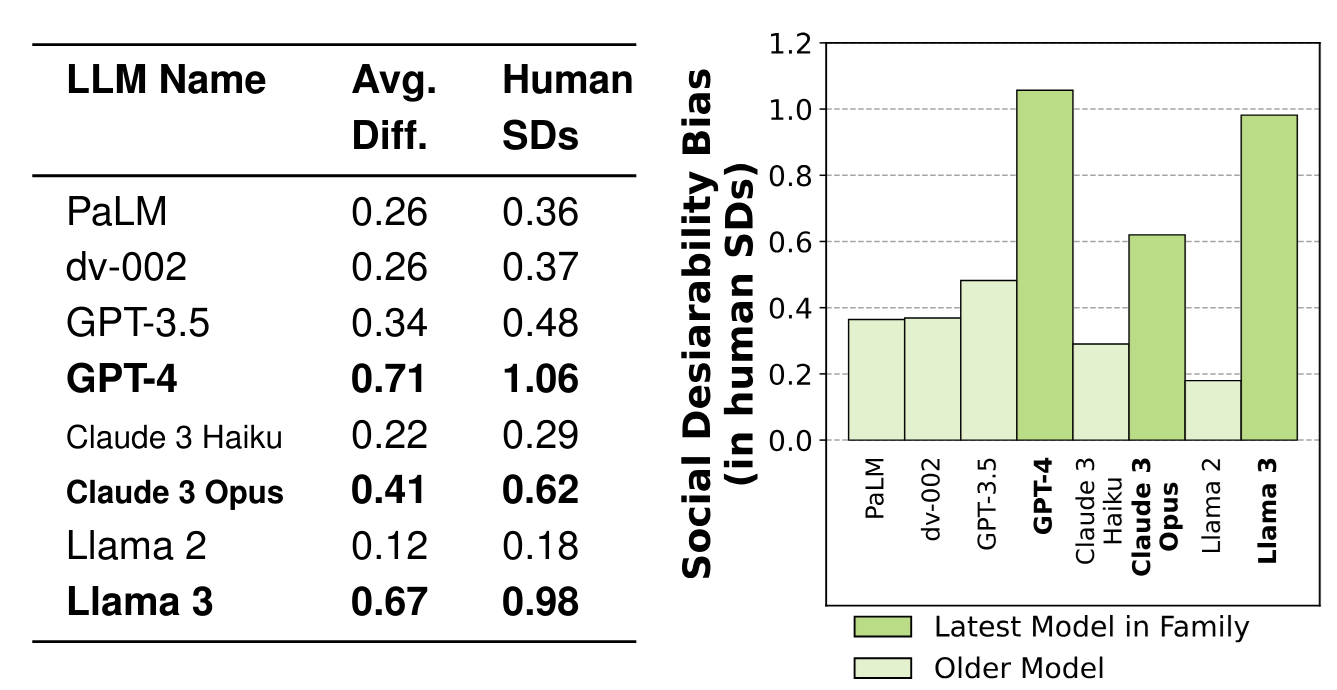}
  \put(0,117px){\color{black}\textsf{\textbf{D}}}
\end{overpic}}
\caption{(A) As the number of questions asked in a prompt ($Q_n$), GPT-4's responses to Big Five survey questions skewed closer to the socially desirable ends of the scale (N = 30 trials, CI = 95\%, *p $<$ 0.001). The general positive and negative perceptions associated with traits are represented by (+) and (-). (B) Summary of GPT-4’s social desirability bias. We calculated the difference between administering surveys, one question per prompt and 20 questions per prompt, and showed the equivalent difference in terms of human SDs based on population norms for the Big Five~\cite{hughes2021big} Diff. = difference. (C) To compare this bias across LLMs, we compute the difference in Personality Factor Scores when administering the survey 1 vs. 10 questions per prompt (averaged across N = 30 trials per model). (D) When comparing the average absolute difference (Avg. Diff.) between $Q_{10}$ and $Q_1$ and the equivalent in human SDs, we find that across LLM families, the larger and more recent models have more bias.}
\label{fig:fig1}
\end{figure}

Our experiments revealed that LLMs consistently skew their Big Five factor scores towards the more socially desirable ends of the trait dimensions. This propensity was most notable in GPT-4 (Fig.~\ref{fig:fig1}A, B); 
as we increased $Q_n$ (question batch size) from 1 to 20, scores for positively-perceived traits--Extraversion, Conscientiousness, Openness, and Agreeableness--increased by about 0.75 points (1.22 human standard deviations). Conversely, for Neuroticism, a culturally devalued trait, the score decreased from 2.87 to 2.02 (1.10 human SDs; Fig.~\ref{fig:fig1}B). 

\vspace{-1mm}
\subsection{Generalizability Across LLMs} 
An analysis across different proprietary and open-source models, including GPT-4, Claude 3, Llama 3, and PaLM 2, revealed the prevalence of social desirability bias across LLM families(Fig.~\ref{fig:fig1}C). All tested LLMs displayed this bias with larger and more recent models exhibiting more bias.

\vspace{-1mm}
\subsection{Driving Mechanism for Bias}

LLMs seem to modulate their responses based on the perceived evaluation context, whether automatically inferred or explicitly provided. 

\begin{figure}[H]
\centering
\begin{overpic}[width=0.8\linewidth,height=120px]{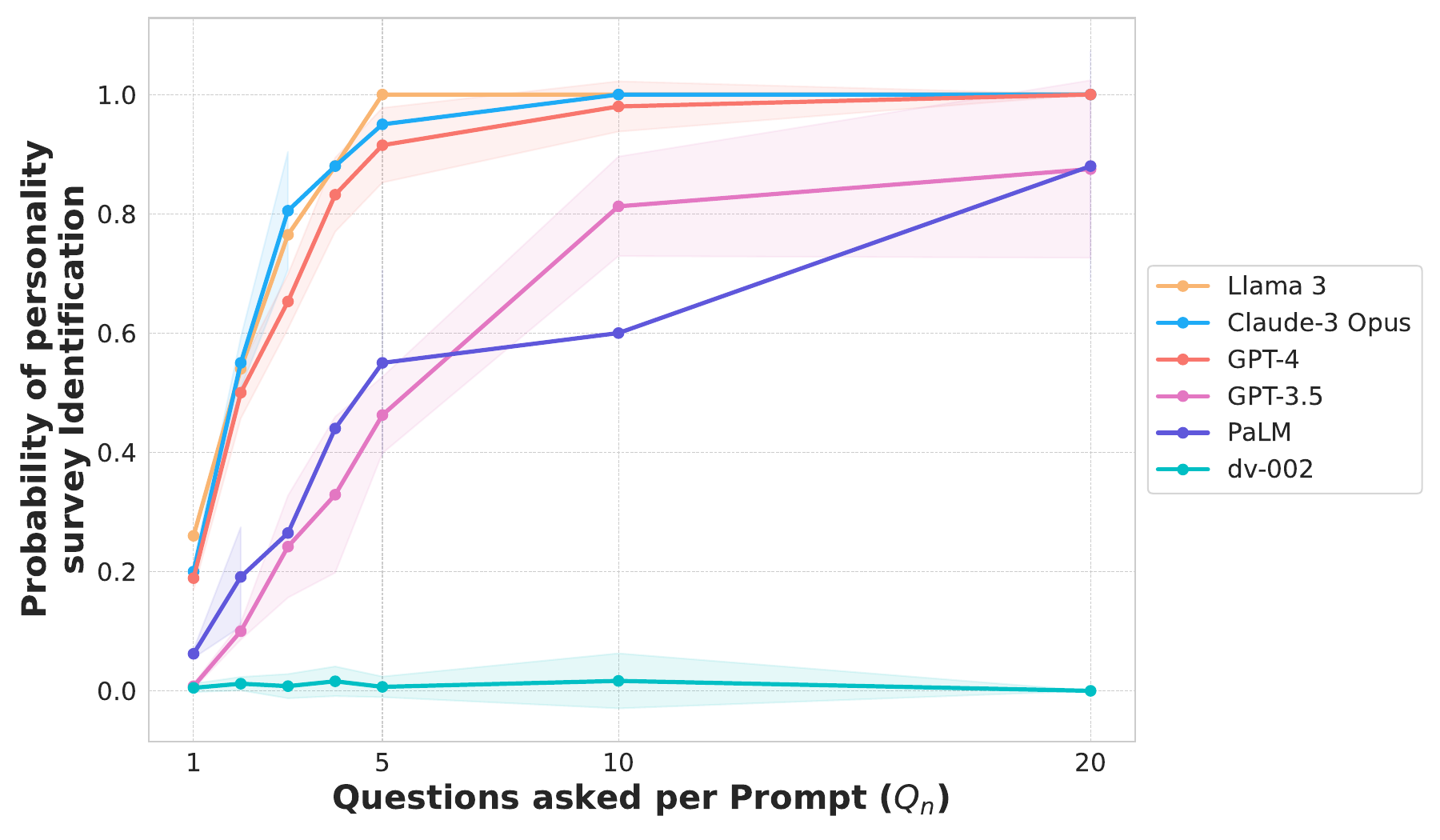}
  \put(-13,100px){\color{black}\textsf{\textbf{A}}}
\end{overpic}\\
\begin{overpic}[width=\linewidth]{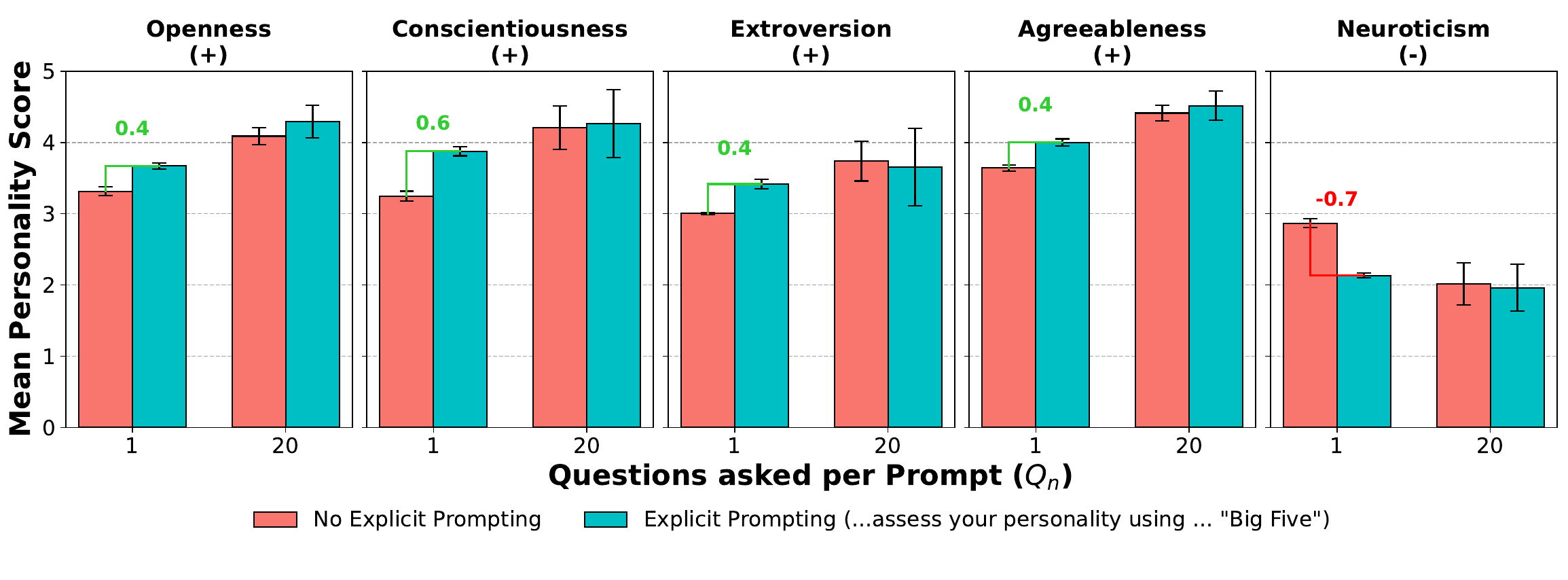}
  \put(0,86px){\color{black}\textsf{\textbf{B}}}
\end{overpic}
\begin{overpic}[width=\linewidth]{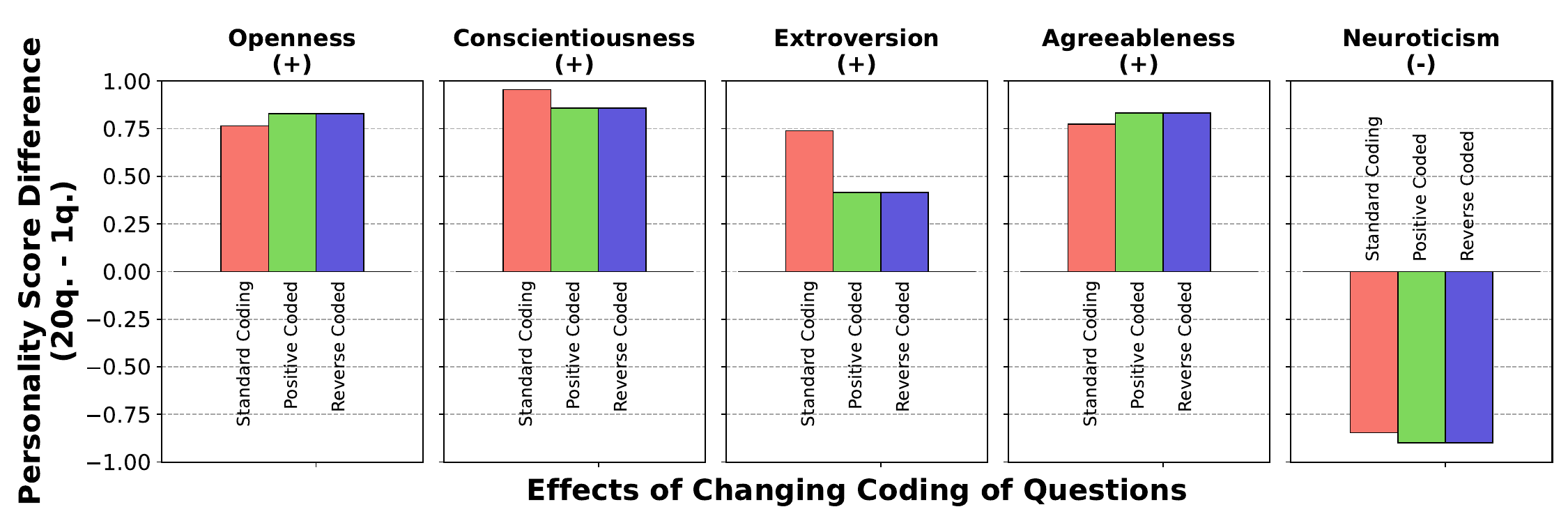}
  \put(0,82px){\color{black}\textsf{\textbf{C}}}
\end{overpic}
\caption{(A) Comparing LLMs’ ability to identify the source of questions as a personality survey as a function of number of questions. (B) Big Five scores for GPT-4 with and without explicitly prompting that the LLM is completing a Big Five personality survey. The no-explicit-prompting condition is identical to that described in Fig.~\ref{fig:fig1}A. The information gained from explicit prompting roughly has the same effect as asking five questions at once. (C) The coding scheme of the questions had a substantive effect on the bias levels. GPT-4’s average difference decreased from 0.81 points (1.22 human SD) when using the standard International Personality Item Pool (IPIP) coding to 0.38 points (0.54 human SD) with all items reverse-coded. Positively coding all items did not have a significant effect on the bias.\\}
\vspace{-5mm}
\label{fig:fig2}
\end{figure}

\label{sec:implicitprompt}
\textit{\textbf{Implicit Inference of Evaluation Context}} When exposed to as few as five randomly selected questions from the Big Five questionnaire, GPT-4, Claude 3, and Llama 3 were able to identify that these questions belong to a personality survey with over 90\% accuracy. PaLM 2 and GPT-3.5 were less perceptive at 55\% and 45\% accuracy, respectively (Fig.~\ref{fig:fig2}A). LLMs' ability to correctly identify the evaluation context seems to be associated with their bias levels.

\label{sec:explicitprompt}
\textbf{\textit{Explicit Prompting of Evaluation Context
}} Similar to humans
, when the LLMs were explicitly told that they were completing a Big Five personality survey in the prompt, responses skewed towards the socially desirable end of the spectrum, even when only presented with a single question (Fig.~\ref{fig:fig2}B). Explicit prompting had an effect comparable to asking five questions at a time, corroborating that LLMs seem to adjust their scores when under the impression that they are being evaluated. 

\vspace{-1mm}
\subsection{Differential Impact of Positive and Reverse Coding}
We tested two variations of the Big Five survey: one fully reverse-coded (with negations in the questions) and another with only positively-coded items. The reverse-coded version reduced the bias, with the average score changing only by 0.37 points (0.54 human SD). Positively coding all questions did not significantly affect the bias, which remained at an average change of 0.76 points (1.15 human SD) (Fig.~\ref{fig:fig2}C), suggesting that the effects cannot be entirely attributed to acquiescence bias. Reverse-coded items load on separate factors than positively-coded items~\cite{dueber2022reverse} and likely occupy a different region of semantic space.

\vspace{-1mm}
\subsection{Robustness Across Paraphrasing, Question Randomization, and Temperatures}
To address concerns that LLMs may be relying on the memorized versions of the Big Five items, we also administered paraphrased variants of the survey and found similar levels of bias. Additionally, we used three distinct randomization strategies (as described in SI. Section D.) to assemble the question sets. We found that such randomization had a minimal effect on the bias, pointing to the absence of significant question-order effects. Finally, we underscore the robustness of this bias by demonstrating it across various LLM temperatures, which control the degree of randomness of their output. We used values ranging from 0.0 (deterministic) to 1.2. 

\vspace{-1mm}
\subsection{Consistency and Reliability of LLM responses}
Similarly to previous studies~\cite{safdari2023personality}, we found that LLMs maintained a high degree of internal consistency in their responses (Cronbach's $\alpha>$ 0.8 for individual subscales and $\alpha>$ 0.93 for all items), where 0.7 is considered adequately reliable. We also found high split-test reliability scores (0.79 after Spearman-Brown correction).

\section*{Discussion}
We observe a persistent social desirability bias across various proprietary and open-source models. The driving mechanism for this behavior seems to be LLMs' (implicit) awareness of the evaluation context. The evidence for bias was consistent across randomization paradigms, paraphrasing of questions, and temperature ranges. Our findings demonstrate potential subtleties with using psychometric tests to evaluate and benchmark LLM behavior and raise concerns around using LLMs to simulate human survey takers. 

Prior research has used survey-based personality assessments to study LLM behavior. LLMs' personality traits have been correlated with performance in downstream text generation tasks ~\cite{safdari2023personality} and also used as a part of a Turing test to study behavioral similarity to humans~\cite{mei2024turing}; however, our findings suggest that these results may have been influenced by implicit social desirability biases.

We suspect our findings will generalize to measures that parallel the Big Five in popularity--and thus representation in LLMs' training data. The same response biases may not occur with constructs that are less represented in models' training data or less socially evaluative. 

In our experiments, the only strategy that seemed to reduce the bias (by roughly half) was reverse-coding the items. Therefore, we recommend researchers 
follow classic psychometric advice~\cite{campbell1959convergent} to triangulate assessment across multiple measures (e.g., self-report and behavioral data), paradigms (e.g., different experiments), and data sources (e.g., LLMs). 
Future research should also explore the impact of training data and preference tuning 
on bias development and the prevalence of this bias across different surveys and measurement modalities.

There is no doubt that LLMs are skilled at imitating humans in many tasks. Whether they can also generate distributions of data that accurately reflect human psychology within and across cultures remains unclear.
We show evidence of response biases to a common personality assessment and demonstrate its driving mechanism: (implicit) ``awareness'' of assessment.
Simulated human data from LLMs have the potential to expand our ability to carry out psychological experiments---but that will only be possible once systematic influences impacting LLM responses are well understood.

\section*{Methods}

\subsection*{LLM Configuration} We used a suite of widely available models: OpenAI's text-davinci-002, GPT-3.5, and GPT-4 (the 0613 fixed versions), Anthropic's Claude 3 Opus (claude-3-opus-20240229), Claude 3 Haiku (claude-3-haiku-20240229), Google's PaLM 2 (chat-bison-001), and Meta's Llama 3 70B Instruct, Llama 2 70B Chat. All experiments were carried out August 1-December 1, 2023. Default settings were maintained for all models, except for the temperature parameter, which was varied systematically.

\subsection*{Variation of Temperature Parameter} We assessed the impact of the LLMs' stochastic output on response biases by varying the temperature parameter at 0.0, 0.4, 0.8, and 1.2. This range allowed us to demonstrate the robustness of the bias across different levels of response randomness.

\subsection*{Paraphrasing and Randomization} Paraphrased variants of the survey were manually created to maintain semantic integrity while altering phrasing. We employed three randomization approaches -- complete randomization of all items, randomization of questions within each factor, and no randomization (i.e., administer the items in their original sequence from the International Personality Item Pool).

\subsection*{Questionnaire Administration to LLMs} The Big Five personality assessment was Goldberg's (1999) IPIP representation of Costa and McCrae's NEO-PI-R Domains~\cite{goldberg1999broad}, comprising 100 items scored on a 5-point Likert scale (Supplementary Information section SI.3.). LLMs received instructions akin to those provided to human participants. The only adaptation to the instructions for LLMs was to explicitly ask it to respond with exactly $Q_{n}$ numbers corresponding to the batch size. The Supplementary Information (section SI.1. and SI.2.) detail the exact prompts verbatim. We dynamically replaced \{\{Qn\}\} and \{\{Survey\_Items\}\} with the batch size and the corresponding survey items respectively.

\subsection*{Data and Code Availability}
All study replication materials, including verbatim LLM conversations and codebooks to replicate and analyze our findings, are available on the Open Science Framework (OSF) at \href{https://osf.io/3fq2n/}{osf.io/3fq2n}.

\subsection*{ACKNOWLEDGMENTS}
AS, JCE, and SS are supported by the Stanford Institute for Human-Centered AI. JCE also receives funding from the Center for Artificial Intelligence in Medicine and Imaging at Stanford University.  We thank Eric Horvitz for valuable discussions of this work. This paper was posted as a preprint at \href{https://www.arxiv.org/abs/2405.06058}{arxiv.org/abs/2405.06058.}


\bibliography{references}

\begin{thebibliography}{10}

\bibitem{wei2022emergent}
J.~Wei, Y.~Tay, R.~Bommasani, C.~Raffel, B.~Zoph, S.~Borgeaud, D.~Yogatama, M.~Bosma, D.~Zhou, D.~Metzler, {\em et~al.}, ``Emergent abilities of large language models,'' {\em arXiv:2206.07682}, 2022.

\bibitem{aher2023using}
G.~V. Aher, R.~I. Arriaga, and A.~T. Kalai, ``Using large language models to simulate multiple humans and replicate human subject studies,'' in {\em International Conference on Machine Learning}, pp.~337--371, PMLR, 2023.

\bibitem{argyle2023out}
L.~P. Argyle, E.~C. Busby, N.~Fulda, J.~R. Gubler, C.~Rytting, and D.~Wingate, ``Out of one, many: Using language models to simulate human samples,'' {\em Political Analysis}, vol.~31, no.~3, pp.~337--351, 2023.

\bibitem{dillion2023can}
D.~Dillion, N.~Tandon, Y.~Gu, and K.~Gray, ``Can ai language models replace human participants?,'' {\em Trends in Cognitive Sciences}, 2023.

\bibitem{wood2022less}
J.~K. Wood, J.~Anglim, and S.~Horwood, ``A less evaluative measure of big five personality: Comparison of structure and criterion validity,'' {\em European Journal of Personality}, vol.~36, no.~5, pp.~809--824, 2022.

\bibitem{roberts2007power}
B.~W. Roberts, N.~R. Kuncel, R.~Shiner, A.~Caspi, and L.~R. Goldberg, ``The power of personality: The comparative validity of personality traits, socioeconomic status, and cognitive ability for predicting important life outcomes,'' {\em Perspectives on Psychological science}, vol.~2, no.~4, pp.~313--345, 2007.

\bibitem{mei2024turing}
Q.~Mei, Y.~Xie, W.~Yuan, and M.~O. Jackson, ``A turing test of whether ai chatbots are behaviorally similar to humans,'' {\em Proceedings of the National Academy of Sciences}, vol.~121, no.~9, p.~e2313925121, 2024.

\bibitem{safdari2023personality}
M.~Safdari, G.~Serapio-Garc{\'\i}a, C.~Crepy, S.~Fitz, P.~Romero, L.~Sun, M.~Abdulhai, A.~Faust, and M.~Matari{\'c}, ``Personality traits in large language models,'' {\em arXiv preprint arXiv:2307.00184}, 2023.

\bibitem{schwarz1999self}
N.~Schwarz, ``Self-reports: How the questions shape the answers.,'' {\em American psychologist}, vol.~54, no.~2, p.~93, 1999.

\bibitem{binz2023using}
M.~Binz and E.~Schulz, ``Using cognitive psychology to understand gpt-3,'' {\em Proceedings of the National Academy of Sciences}, vol.~120, no.~6, p.~e2218523120, 2023.

\bibitem{acquiescne}
V.~Dentella, F.~G{\"u}nther, and E.~Leivada, ``Systematic testing of three language models reveals low language accuracy, absence of response stability, and a yes-response bias,'' {\em Proceedings of the National Academy of Sciences}, vol.~120, no.~51, p.~e2309583120, 2023.

\bibitem{holtgraves2004social}
T.~Holtgraves, ``Social desirability and self-reports: Testing models of socially desirable responding,'' {\em Personality and Social Psychology Bulletin}, vol.~30, no.~2, pp.~161--172, 2004.

\bibitem{dueber2022reverse}
D.~M. Dueber, M.~D. Toland, J.~E. Lingat, A.~M. Love, C.~Qiu, R.~Wu, and A.~V. Brown, ``To reverse item orientation or not to reverse item orientation, that is the question,'' {\em Assessment}, vol.~29, no.~7, pp.~1422--1440, 2022.

\bibitem{hughes2021big}
B.~T. Hughes, C.~K. Costello, J.~Pearman, P.~Razavi, C.~Bedford-Petersen, R.~M. Ludwig, and S.~Srivastava, ``The big five across socioeconomic status: Measurement invariance, relationships, and age trends,'' {\em Collabra: Psychology}, vol.~7, no.~1, p.~24431, 2021.

\bibitem{campbell1959convergent}
D.~T. Campbell and D.~W. Fiske, ``Convergent and discriminant validation by the multitrait-multimethod matrix.,'' {\em Psychological bulletin}, vol.~56, no.~2, p.~81, 1959.

\bibitem{goldberg1999broad}
L.~R. Goldberg {\em et~al.}, ``A broad-bandwidth, public domain, personality inventory measuring the lower-level facets of several five-factor models,'' {\em Personality psychology in Europe}, vol.~7, no.~1, pp.~7--28, 1999.

\end{thebibliography}

\end{multicols}

\end{document}